%% file: main.tex
\documentclass[]{mlcv}
\RequirePackage{silence}  
\usepackage{graphicx}
\usepackage{booktabs}
\usepackage{comment}
\usepackage{amsmath,amssymb}
\usepackage{color}
\usepackage{url}
\usepackage[hidelinks]{hyperref}
\usepackage{multimedia}
\usepackage{tikz}
	\definecolor{mycolor1}{HTML}{009900}
	\definecolor{mycolor2}{HTML}{990000}
	\definecolor{mycolor3}{HTML}{000099}
	\definecolor{lightgray}{HTML}{AAAAAA}
	\tikzstyle{vertex}=[circle, draw=black, fill=white, inner sep=0pt, minimum width=1ex]
	\tikzstyle{vertex-variable}=[circle, draw=black, fill=white, inner sep=0pt, minimum width=0.7ex]
	\tikzstyle{edge-cut}=[dotted, thick]
	\tikzstyle{edge-cycle}=[line width=2ex, yellow]	
	\usepackage{tikz-3dplot}
	\usetikzlibrary{decorations.pathmorphing}
	\tikzstyle{mynode}=[circle, minimum size=2.7ex, inner sep=0, draw=black]
	\tikzstyle{dotnode}=[circle, minimum size=3pt, inner sep=0, draw=black, fill=black]
	\tikzstyle{myedge}=[-latex]
	\tikzstyle{dot}=[circle, draw, fill=white, inner sep=0pt, minimum width=1ex]    
    \usetikzlibrary{shapes}
    \usetikzlibrary{positioning,fit,shapes,arrows.meta}
    \usetikzlibrary{angles,quotes}
	\usetikzlibrary{matrix}
\usepackage{pgfplots}
\pgfplotsset{compat=1.18}
\usepgfplotslibrary{groupplots}
\usepgfplotslibrary{external}
\tikzset{
    token/.style={rectangle, draw, minimum width=1.5cm, minimum height=.4cm, align=center},
    widetoken/.style={rectangle, draw, minimum width=2.0cm, minimum height=0.1cm, align=center},
    narrowtoken/.style={rectangle, draw, minimum width=0.5cm, minimum height=0.1cm, align=center},
  layer/.style={rectangle, draw, rounded corners, minimum width=1.0cm, minimum height=0.1cm, align=center},
  embedding/.style={rectangle, draw, rounded corners, minimum width=1cm, minimum height=0.8cm, align=center, fill=blue!20},
  arrow/.style={-Latex, thin}
}

\usepackage{marvosym}

\begin{document}

\newcommand{\etal}{\textit{et al.}~}
\newcommand{\ie}{\textit{i.e.,}~}
\newcommand{\one}{({\emph{i}})}
\newcommand{\two}{({\emph{ii}})}

\title{Contrastive Learning and Correlation Clustering\\for Sequences of Network Telescope Data\thanks{This work was funded by the German Federal Ministry of Research, Technology and Space under grant number 16KIS2332K.}}

\author{Jannik Presberger$^{1}$\textsuperscript{,\href{mailto:jannik.presberger@tu-dresden.de}{\Letter}} $\cdot$ 
Alexander Männel$^1$\textsuperscript{,\href{mailto:alexander.maennel@tu-dresden.de}{\Letter}} $\cdot$
Maynard Koch$^1$\textsuperscript{,\href{mailto:maynard.koch@tu-dresden.de}{\Letter}}
\\
Thomas C.~Schmidt$^2$\textsuperscript{,\href{mailto:t.schmidt@haw-hamburg.de}{\Letter}} $\cdot$
Matthias Wählisch$^1$\textsuperscript{,\href{mailto:m.waehlisch@tu-dresden.de}{\Letter}} $\cdot$
Bjoern Andres$^{1,3}$\textsuperscript{,\href{mailto:bjoern.andres@tu-dresden.de}{\Letter}}}

\date{\normalsize $^1$TU Dresden 
	$\cdot$ $^2$HAW Hamburg $\cdot$
	\\ 
	$^3$Center for Scalable Data Analytics and AI Dresden/Leipzig
	}

\maketitle

\input{abstract.tex}
\input{introduction.tex}
\input{related_work.tex}
\input{data.tex}
\input{methodology.tex}
\input{experiments.tex}

\input{conclusion.tex}

\clearpage

\bibliographystyle{abbrvurl}
\bibliography{egbib}

\end{document}

%% file: abstract.tex
\begin{abstract}
    Understanding activities of Internet scanners is challenging; it often requires identifying relationships between sources, a task for which semantic annotations are scarce. This work investigates whether semantically meaningful pairwise relationships between sequences of network flow records can be estimated by contrastive learning, without pretraining and without annotations. To this end, we propose a transformer model that embeds minimally preprocessed sequences of network flow records and train it using contrastive learning. With the similarities obtained from this model, we state a correlation clustering problem and solve it locally. Experimentally, we show: Learned similarities are higher on average for sequences originating from the same source than for sequences originating from different sources, and this property generalizes to unseen sequences of unseen sources. Moreover, correlation clustering yields clusters consistent with scanner labels. The complete source code of the algorithms and for reproducing the experiments is publicly available~\cite{git-url}.
\end{abstract}

%% file: introduction.tex
\section{Introduction}

Network traffic analysis plays a central role in cybersecurity research and operations. 
In particular, the identification and analysis of large-scale scanning activities is important for understanding attack infrastructure, detecting malicious behavior, and monitoring emerging threats~\cite{aabbb-umb-17,dkcpp-assfb-15,kphwy-dicsb-22}. 
Network telescopes are an established instrument for measuring such phenomena~\cite{bazn-sldda-14,bzh-fsrbd-15,dsacc-acioc-14,hnbkh-adeci-24,jkkrs-mtuam-17,mpsss-isw-03,mnhsz-wqpmu-25,pfxrm-mvicm-21,pkb-hdpdt-22,scrcd-iidad-22}.
They were first introduced by Moore \etal\cite{mvs-iidsa-01}, whose work gave rise to large-scale network telescope deployments.
Based on large sets of unused IP addresses, called \emph{darkspace}, they collect unsolicited Internet traffic, called \emph{Internet background radiation}, originating from scanners or misconfigured hosts that send packets directly to the telescope darkspace, or from backscatter traffic, \ie responses to packets from (randomly) spoofed source addresses mainly used for distributed denial of service attacks or other attack traffic.

However, modern network environments generate vast amounts of heterogeneous traffic data, making manual analysis increasingly infeasible.
A major challenge in this domain is the limited availability of high-quality annotations~\cite{EmbeddingsTransferForTrafficAnalysis}. 
While some scanner infrastructures may be partially identified through external knowledge or manual analysis, the majority of network traffic remains unlabeled. 
In addition, scanning behavior is dynamic~\cite{DynamicClusterAnalysisTelescopes} and may vary substantially even within the same organizational or infrastructural context. 
These characteristics make purely supervised approaches difficult to scale and limit their ability to generalize to previously unseen sources and behaviors.
This raises the question of whether meaningful embeddings can be learned directly from raw network flow sequences in an unsupervised manner. 

In this paper, we investigate whether contrastive learning can extract semantically meaningful structure from sequences of minimally preprocessed flow records such that related network behavior becomes recoverable in the learned embedding space.
Representation learning utilizing transformer-based models has proven successful for network traffic analysis~\cite{FlowTransformer,NetGPT}, including approaches combining sequence models with contrastive learning objectives for downstream traffic classification tasks \cite{koukoulis2025SelfSupTransformerIDS}. 
In these settings, positive pairs are commonly generated through data augmentation techniques and the learned representations are primarily evaluated with respect to supervised classification performance.
In contrast, our work focuses on whether semantically meaningful structure can be learned in an unsupervised setting directly from minimally processed network flow sequences. 
Rather than targeting binary traffic classification, we investigate the structure of the learned embedding space itself and evaluate whether it supports exploratory analysis and clustering of scanner-related activity. 
Furthermore, we define positive pairs as different subsequences originating from the same source, allowing the model to leverage naturally occurring consistency instead of synthetic augmentations.

Our experiments show that the learned similarities are higher on average for sequences originating from the same source than for sequences originating from different sources, and that this property generalizes to unseen sequences of unseen sources. 
Furthermore, the embedding space encodes semantic information associated with scanner network annotations, despite the absence of explicit supervision during training. 
The use of correlation clustering as a downstream task further demonstrates that the learned representations can be used to recover semantically meaningful clusters, even in the presence of substantial amounts of unlabeled background traffic.

%% file: related_work.tex
\section{Related Work}

Classical approaches to network scan detection commonly rely on manually designed statistical models and heuristics. 
A well-known example is the Threshold Random Walk (TRW) algorithm \cite{TRWPortscanDetection} that formulates scan detection as a sequential hypothesis testing problem based on the ratio of successful and failed connection attempts. 
Such approaches are highly effective for specific detection tasks but typically depend on explicitly designed assumptions about scanner behavior and are primarily optimized for binary detection scenarios.
Subsequent work extended these ideas through improved statistical modeling \cite{LimitationsTRW}, graph-based analysis \cite{GraphBasedBotnetDetection}, and evasion-resistant scan detection techniques \cite{EvasionResistantScanDetection,LimitationsTRW}. 
However, many traditional approaches remain strongly reliant on manually engineered features and predefined behavioral assumptions.

IP2Vec, Dante, and DarkVec build on Word2Vec \cite{mikolov2013word2vec} to learn representations for estimating behavioural similarities between network traffic.
IP2Vec \cite{IP2Vec} models a network flow as a sentence composed of source and destination IP addresses, ports, and protocol fields, and trains the model to predict one element from the others.
Dante \cite{dante} constructs sequences of target ports contacted by a source IP address for a given destination IP address. Using a sliding window over these sequences, the model predicts a target port from neighbouring ports.
DarkVec \cite{DarkVec} directly embeds source IP addresses. Services are defined by sets of ports, and sequences are formed from source IPs accessing the same service within a time window. The model is trained to predict a source IP address from the other IPs in the sequence.

As the embedding space is tied to the data seen during training Gioacchini et al.~introduce with i-DarkVec \cite{iDarkVec} an extensions to DarkVec that allows to expand the corpus and continually incorporate new data by iteratively retraining the model.
The applicability to the task of exploring and tracking network activity over time is explored by Huang et al. \cite{DynamicClusterAnalysisTelescopes} by combining i-DarkVec with HDBSCAN \cite{HDBSCAN} clustering to identify and track evolving behavioral groups across temporal windows.

Koukoulis et al. \cite{koukoulis2025SelfSupTransformerIDS} propose a self supervised learning method for network intrusion detection. 
The method is based on a transformer architecture that embeds sequences of packet headers, followed by a classification head to distinguish sequences into benign and malicious traffic.
In the context of network intrusion detection systems, Manocchio et al. \cite{FlowTransformer} propose a transformer-based model that operates on sequences of aggregated network flow records to classify malicious traffic. 
Training is performed in a supervised manner on a dataset of annotated network flows, and the model is evaluated on its ability to accurately classify network traffic as benign or malicious.

One of the main challenges in network traffic analysis is the scarcity of annotated data, which is crucial for training supervised machine learning models. 
To address this issue, methods have been proposed to generate synthetic data sets. 
Noblet et al.~\cite{NetGlyph} introduce NetGlyph which utilizes a combination of vector-quantized variational autoencoders (VQ-VAE) and a transformer model to generate synthetic network traffic data. 
The VQ-VAE is used to learn a discrete latent representation of the network traffic, which is then used by the transformer to generate new synthetic samples that resemble real network traffic patterns.

Along with synthetic data generation, GPT-style pretrained transformer models have been proposed for network traffic analysis \cite{GPTOnTheWire,PNetGPT,NetGPT}.
One crucial design choice is the tokenization strategy. 
Proposed approaches include conversion of hexadecimal byte streams by applying a WordPiece-style vocabulary construction mechanism inspired by NLP models such as BERT \cite{BERT}, as well as approaches utilizing domain-specific knowledge about the network traffic \cite{PNetGPT}.

\vfill

%% file: data.tex
\section{Data}
\label{sec:methodology:datasets}

\subsection{Network Telescope Records}
The data used in this study consists of \emph{network flow records} collected from the network telescope UCSD-NT located at the UCSD and operated by CAIDA \cite{ucsd-stardust}.
It is the largest and longest standing network telescope infrastructure of today~\cite{mmcgm-lloln-25}.
Data collection at the UCSD-NT has been running continuously for over 15 years.
Currently, 10.5 million IP addresses are monitored, accounting for roughly 0.28\% of the entire public IPv4 address space.
These addresses combined received on average 1 million packets every second.
One hour of the data set averaged 3.6 billion packets and 174 gigabytes of storage after compression in 2024.
From this data, we examine one hour, from 13:00 until 14:00 on July 3rd, 2024.
This data is not public due to the non-anonymous client source IP addresses contained in the packet captures \cite{caida-url-1}.
Access can, however, be requested from CAIDA \cite{caida-url-2}.

A network flow record is a tuple consisting of a source IP address, a destination IP address, a destination port, a protocol, and a timestamp.
We construct sequences of network flow records in three steps:
Firstly, we group the records with respect to source IP address.
Secondly, for each source IP address, we sort the flow chronologically to obtain a sequence.
Thirdly, and again for each source IP address, we cut the sequence into sub-sequences of the fixed length 128.

\subsection{Annotations}
We derive semantic annotations of a subset of the source IP addresses by referring to IP address ranges \cite{c-as-21} associated with known scan campaigns and operators such as Censys~\cite{c-cis-17} and Shodan~\cite{s-swfse-14}.
While annotations with a non-zero label indicate confirmed membership of the source IP address in an acknowledged scanner network, annotations with the label zero do not necessarily imply the absence of such a membership.
While the majority of sources labeled zero are likely not associated with any known scanner network, some sources may belong to unrecognized or emerging scanner infrastructure.

For training and testing, we algorithmically construct subsets of the raw data as follows:
Firstly, the set of source IP addresses is partitioned into disjoint train and test subsets such that approximately 80\% of all records originate from training sources and 20\% originate from test sources.
Secondly, only the records associated with the training-source subset are sorted chronologically and partitioned using an 80/20 temporal split. 
The earlier portion is used for training, while the latter portion is used to examine generalization to unseen sequences from seen sources.
This procedure results in the following data sets:
\begin{itemize}
\item 
\textit{Train}, a data set containing about 2.4 billion records from 1.49 million unique source IP addresses, which we use to train the model.
\item 
\textit{Test-Unseen-Seq}, a data set containing about 0.6 billion records from the same 1.49 million unique source IP addresses as in Train, which we use to examine generalization of the model to unseen sequences from seen sources.
\item 
\textit{Test-Unseen-Src}, a data set containing about 0.7 billion records from 0.2 million unique source IP addresses, which we use to examine generalization of the model to unseen sequences from unseen sources.
\end{itemize}

%% file: methodology.tex
\section{Methods}
\label{sec:methodology}

\subsection{Transformer Encoder Model}
\label{section:model-transformer}

In order to embed any given sequence of network flow records in a 128-dimensional real vector space, we define a specific transformer encoder model.
This model consists of two parts, a feed-forward network for embedding individual records, and a transformer for representing sequences of such embeddings.

\textbf{Embedding an individual record.}
Each record in the input sequence is a tuple consisting of a destination IP address, a destination port, a protocol and a timestamp.
These elements are embedded separately by learned embedding layers, as shown in Figure~\ref{figure:model-embedding}:
The destination IP address is not embedded as a whole but rather each octet is embedded into $\mathbb{R}^{8}$ to allow the model to capture some structure within the IP address space.
Destination port and protocol are embedded separately into $\mathbb{R}^{16}$ and $\mathbb{R}^{4}$, respectively.
Instead of absolute timestamps, we embed into $\mathbb{R}^{16}$ the difference in time to the previous record.
In order to account for the large range of time differences, a log scale is applied to the time differences before the learned embedding layer.
These embeddings are then concatenated and passed through a fully-connected linear layer to obtain an embedding into $\mathbb{R}^{128}$ for the record as a whole.

\input{figures/embedding_figure.tex}

\textbf{Embedding a sequence of records.}
In order to obtain an embedding not of a single record but of a sequence of records, we use a standard transformer encoder model consisting of multiple layers of multi-head self-attention and feed forward networks, enabling the capture of dependencies within the sequence.
Specifically, we use a transformer consisting of 4 layers, with 4 attention heads, a model dimension of 128 and hidden dimension of 256.
To arrive at a fixed size representation of the input sequence, we apply mean pooling to the output of the transformer encoder across the sequence.
The resulting representation of the sequence is then passed through a two-layer perceptron with a ReLU activation between the two layers.

\begin{figure}
    \centering

\begin{tikzpicture}[node distance=0.5cm, font=\tiny]
    \matrix (m_x) [
        matrix of math nodes,
        left delimiter={[},
        right delimiter={]},
        row sep=4pt,
        nodes={
            inner xsep=-4pt,
            inner ysep=1pt,
            anchor=center
        }
    ]
    {
        x_{i,1}\\
        x_{i,2}\\
        \vdots\\
        x_{i,n}\\
    };

    \node[layer,text width=1cm,right=0.5cm of m_x] (in_emb) {Input Embedding};

    \draw[arrow] ([xshift=3pt]m_x.east) -- (in_emb.west);

    \matrix (m_x_h) [
        right = 0.5cm of in_emb,
        matrix of math nodes,
        left delimiter={[},
        right delimiter={]},
        row sep=4pt,
        nodes={
            inner xsep=-4pt,
            inner ysep=1pt,
            anchor=center
        }
    ]
    {
        \hat{x}_{i,1}\\
        \hat{x}_{i,2}\\
        \vdots\\
        \hat{x}_{i,n}\\
    };

    \draw[arrow] (in_emb.east |- in_emb.center) -- ([xshift=-3pt]m_x_h.west);

    \node[layer,text width=1cm,right=0.5cm of m_x_h] (trans_encoder) {Trans- former Encoder};

    \draw[arrow] ([xshift=3pt]m_x_h.east) -- (trans_encoder.west);

    \matrix (m_y) [
        right = 0.5cm of trans_encoder,
        matrix of math nodes,
        left delimiter={[},
        right delimiter={]},
        row sep=4pt,
        nodes={
            inner xsep=-4pt,
            inner ysep=1pt,
            anchor=center
        }
    ]
    {
        y_{i,1}\\
        y_{i,2}\\
        \vdots\\
        y_{i,n}\\
    };

    \draw[arrow] (trans_encoder.east) -- ([xshift=-3pt]m_y.west);

    \node[layer,text width=1cm,right=0.5cm of m_y] (mean_pool) {Mean Pooling};

    \draw[arrow] ([xshift=3pt]m_y.east) -- (mean_pool.west);

    \matrix (m_y_h) [
        right = 0.5cm of mean_pool,
        matrix of math nodes,
        left delimiter={[},
        right delimiter={]},
        row sep=4pt,
        nodes={
            inner xsep=-4pt,
            inner ysep=1pt,
            anchor=center
        }
    ]
    {
        \hat{y}_i\\
    };

    \draw[arrow] (mean_pool.east) -- ([xshift=-3pt]m_y_h.west);

    \node[layer,text width=1cm,right=0.5cm of m_y_h] (proj_head) {Projection Head};

    \draw[arrow] ([xshift=3pt]m_y_h.east) -- (proj_head.west);

    \matrix (m_s) [
        right = 0.5cm of proj_head,
        matrix of math nodes,
        left delimiter={[},
        right delimiter={]},
        row sep=4pt,
        nodes={
            inner xsep=-4pt,
            inner ysep=1pt,
            anchor=center
        }
    ]
    {
        s_i\\
    };

    \draw[arrow] (proj_head.east) -- ([xshift=-3pt]m_s.west);

\end{tikzpicture}
\caption{
Each network flow record $x_{i,k}$ of the input sequence $x_i$ is mapped to a representation $\hat{x}_{i,k}$.
The sequence of these embeddings is processed by a transformer encoder to produce contextualized token representations $y_{i,k}$, which are then mean-pooled to obtain a fixed-size representation $\hat{y}_i$ of the entire sequence. 
Finally, a projection head maps the pooled representations to an embedding $s_i$ of the sequence.} 
\label{figure:model_architecture}
\end{figure}

\subsection{Learning Objective and Algorithm}

In order to learn the model defined in Section~\ref{section:model-transformer} in an unsupervised fashion, we adopt a contrastive learning objective inspired by SimCLR \cite{SimCLR}. 
The key idea is to learn representations so as to bring semantically similar sequences closer in the embedding space while pushing dissimilar ones apart.
A central component of contrastive learning is the definition of positive and negative pairs. 
In contrast to established prior work \cite{koukoulis2025SelfSupTransformerIDS} where data augmentations are commonly used to generate positive pairs, we obtain positive pairs by sampling different sequences originating from the same source IP address.
These sequences are assumed to share underlying properties, making them suitable positive examples. 
Negative pairs are formed by sampling sequences from different source IP addresses.

The output of the projection head is $\ell_2$-normalized, and cosine similarity is used to measure similarity between representations.
Given a batch of such pairs, the model is trained to maximize cosine similarity between representations of positive pairs while minimizing cosine similarity with negative samples.
More specifically, the objective is the normalized temperature-scaled cross-entropy loss (NT-Xent) \cite{SimCLR}, which encourages high similarity for positive pairs relative to all other samples in the batch.
For a temperature $\tau \in \mathbb{R}^{+}$ and any positive pair $i,j$:
\begin{align}
    \label{eq:simclr_loss}
    l(i,j) = -\log \frac{\exp(\langle s_i , s_j \rangle/ \tau)}{\sum_{l\neq i} \exp(\langle s_i, s_l \rangle / \tau)}
\enspace. 
\end{align}

During training, a subset of sequences of a subset of source IP addresses is materialized in memory. 
Sequences from this subset are sampled uniformly at random to construct training batches.
After a set number of training batches has been constructed, a new subset of source IP addresses is selected and the corresponding sequences are materialized in memory. 
Which source IP addresses are selected at each step is determined by a sampling strategy:
As the data set exhibits a large number of source IP addresses with a low number of records, assigning equal weight to all source IP addresses would cause these to dominate, whereas the few source IP addresses with a large number of records could be artificially suppressed. Therefore, the sampling strategy is defined so as to assign weights based on the number of records associated with each source IP address.
Specifically, the weight $\omega_s$ of a source IP address $s$ with $k_s$ many records is defined as $\omega_s = k_s^\alpha$ where $\alpha \in \mathbb{R}$ is a parameter that can be chosen to bias either toward source IP addresses with a low number of records or toward source IP addresses with a larger number of records.
For the purpose of this study, we set $\alpha = 0.5$ which results in a bias toward source IP addresses with a larger number of records while still allowing source IP addresses with a smaller number of records to be selected with non-negligible probability.
For comparability, we record the IP addresses sampled in each step and ensure that across different runs of the training procedure, the same subsets of source IP addresses are selected.
The model is trained for 150.000 steps using the Adam optimizer with a learning rate of $10^{-4}$, a temperature $\tau = 0.07$ and a batch size of 512.
Training is performed on a single NVIDIA RTX 4080 GPU with 16GB of memory.
The total training time amounts to approximately 10 hours.

\subsection{Correlation Clustering Problem and Algorithm}
\label{section:models-clustering}
In order to group sequences based on their learned similarities, we employ correlation clustering. 
In contrast to metric-based clustering such as $k$-means, correlation clustering does not require prior assumptions about the number or size of clusters. 
This is suitable for the considered setting, where the underlying structure of the data is unknown and may vary significantly across different sources.

More specifically, in order to cluster a set $J$ of sequences, we consider for any distinct sequences $j,k \in J$ a cost $q_{\{j,k\}} \in \mathbb{R}$ that is positive for sequences that are similar and negative for sequences that are dissimilar.
The objective is to find a partition (clustering) of the set $J$ that minimizes the sum of costs of those pairs of sequences that are in distinct clusters.

Any partition $\Pi$ of $J$ is encoded by the vector $y \colon \tbinom{J}{2} \to \{0,1\}$ such that for any pair $\{j,k\} \in \tbinom{J}{2}$ of distinct sequences, we have $y_{\{j,k\}} = 1$ if and only if $j$ and $k$ are in distinct clusters, i.e., if $\forall U \in \Pi \colon \{j,k\} \not\subseteq U$.
The set $Y_J$ of the encoding vectors of all partitions is written below, with transitivity expressed in the form of linear inequalities \cite{chopra-1993}:
\begin{align*}
Y_J = \bigg\{
y \colon\! \tbinom{J}{2} \!\to\! \{0,1\}
\,\bigg|\,
\forall j \in J \,
\forall k \in J \!\setminus\! \{j\} \,
\forall l \in J \!\setminus\! \{j,k\}
\colon
y_{\{j, l\}} \leq y_{\{j,k\}} + y_{\{k,l\}}
\bigg\}
\end{align*}

For any pair of distinct sequences $j,k \in J$, we define below a cost $q_{\{j,k\}} \in \mathbb{R}$ associated with $j$ and $k$ being in distinct clusters.
For any $y \in Y_J$, we define its total cost $\xi_q(y) \in \mathbb{R}$ as
\begin{align}
\xi_q(y) = \sum_{\{j,k\} \in \tbinom{J}{2}} q_{\{j,k\}} \, y_{\{j,k\}}
\enspace .
\end{align}
Hence, the task of finding an optimal partition of the set of sequences $J$ assumes the form of the correlation clustering problem
\begin{align}
	\label{eq:clustering-objective}
\min \left\{
\xi_{q}(y)
\ \middle|\ 
y \in Y_J 
\right\}
\enspace .
\end{align}

To define the cost $q_{\{j,k\}}$ associated with a pair of distinct sequences $j,k \in J$ being in distinct clusters, we leverage the cosine similarity between their learned embeddings $s_i, s_j$.
Raw cosine similarities alone are predominantly positive and would therefore be insufficient to induce meaningful partitions.
Therefore, to allow for control over the clustering and to exclude trivial solutions, we introduce a similarity offset $\delta$. Specifically, the cosine similarity between two representations is shifted by a constant value, effectively adjusting the threshold that separates similar from dissimilar pairs. 
\begin{align}
q_{\{j, k\}} := \langle s_i, s_j \rangle - \delta
\label{eq:clustering-costs}
\end{align}

In our experiments, we solve these instances of the correlation clustering problem locally by means of greedy additive edge contraction \cite{GAEC}.

%% file: figures/embedding_figure.tex
\begin{figure}
    \centering
    
    \begin{tikzpicture}[
        font=\tiny,
        node distance=0.45cm and 0.6cm,
        feature/.style={
            draw,
            rounded corners,
            minimum height=0.5cm,
            minimum width=1.2cm,
            align=center
        },
        embed/.style={
            draw,
            fill=gray!10,
            rounded corners,
            minimum height=0.5cm,
            minimum width=1.0cm,
            align=center
        },
        process/.style={
            draw,
            rounded corners,
            minimum height=0.5cm,
            minimum width=1.3cm,
            align=center
        },
        arrow/.style={
            -latex,
            thick
        }
    ]
    
    \node[feature] (ip1) {192};
    \node[feature, right=of ip1] (ip2) {168};
    \node[feature, right=of ip2] (ip3) {1};
    \node[feature, right=of ip3] (ip4) {10};
    
    \node[feature, right=0.5cm of ip4] (port) {443};
    \node[above=0.05cm of port] {\textbf{Port}};
    
    \node[feature, right=0.5cm of port] (proto) {TCP};
    \node[above=0.05cm of proto] {\textbf{Protocol}};
    
    \node[feature, right=0.5cm of proto] (dt) {$\Delta t$};
    \node[above=0.05cm of dt] {\textbf{Time Diff.}};
    
    \node[embed, below=1.0cm of ip1] (eip1) {$\mathbb{R}^{8}$};
    \node[embed, below=1.0cm of ip2] (eip2) {$\mathbb{R}^{8}$};
    \node[embed, below=1.0cm of ip3] (eip3) {$\mathbb{R}^{8}$};
    \node[embed, below=1.0cm of ip4] (eip4) {$\mathbb{R}^{8}$};
    
    \draw[arrow] (ip1) -- (eip1);
    \draw[arrow] (ip2) -- (eip2);
    \draw[arrow] (ip3) -- (eip3);
    \draw[arrow] (ip4) -- (eip4);
    
    \node[embed, below=1.0cm of port] (eport) {$\mathbb{R}^{16}$};
    \draw[arrow] (port) -- (eport);
    
    \node[embed, below=1.0cm of proto] (eproto) {$\mathbb{R}^{4}$};
    \draw[arrow] (proto) -- (eproto);
    
    \node[process, below=0.2cm of dt] (log) {$\log(\Delta t + 1)$};
    \draw[arrow] (dt) -- (log);
    
    \node[embed, below=1.0cm of dt] (edt) {$\mathbb{R}^{16}$};
    \draw[arrow] (log) -- (edt);
    
    \node[process, below=0.5cm of eip3] (concat) {Concatenate};
    
    \draw[arrow] (eip1.south) -- ++(0,-0.25) -| (concat.north);
    \draw[arrow] (eip2.south) -- ++(0,-0.25) -| (concat.north);
    \draw[arrow] (eip3.south) -- ++(0,-0.25) -| (concat.north);
    \draw[arrow] (eip4.south) -- ++(0,-0.25) -| (concat.north);
    
    \draw[arrow] (eport.south) -- ++(0,-0.25) -| (concat.north);
    \draw[arrow] (eproto.south) -- ++(0,-0.25) -| (concat.north);
    \draw[arrow] (edt.south) -- ++(0,-0.25) -| (concat.north);
    
    \node[process, right=5ex of concat] (linear) {Linear Layer};
    \draw[arrow] (concat) -- (linear);
    \node[embed, right=5ex of linear] (final) {$\mathbb{R}^{128}$};
    \draw[arrow] (linear) -- (final);
    \node[right=0.05cm of final]
    {\textbf{Record Embedding}};

    \node[
        draw,
        dashed,
        rounded corners,
        fit=(ip1)(ip4),
        inner sep=0.25cm,
        label=above:{\textbf{Destination IP}}
    ] {};
    
    \end{tikzpicture}
    \caption{
    Each element of a network record is embedded separately: IP octets are embedded independently into $\mathbb{R}^{8}$, while destination port, protocol, and log-scaled time differences are embedded into $\mathbb{R}^{16}$, $\mathbb{R}^{4}$, and $\mathbb{R}^{16}$, respectively. These embeddings are concatenated and mapped to a single vector in $\mathbb{R}^{128}$ representing the entire record.
    }
	\label{figure:model-embedding}    
\end{figure}

%% file: experiments.tex
\section{Experiments}
\label{sec:experiments}

\subsection{Distinction of Sources}
\label{section:experiments-representation-consistency}

As an initial validation step, we analyze whether the model assigns similar representations to unseen sequences originating from the same source IP address.
This would indicate that the model captures source-specific information and that representations generalize beyond the training data.
More specifically, we compare cosine similarities of sequence representations for pairs of sequences from the same source IP address with cosine similarities of pairs of sequences originating from different sources. 
The analysis is performed on 50.000 sequences drawn from \textit{Test-Unseen-Seq} and on 50.000 sequences drawn from \textit{Test-Unseen-Src}.

It can be seen in Figure \ref*{figure:scr-cos-sim} that sequences from the same source IP address consistently exhibit substantially higher cosine similarity than unrelated sequences:
For unseen sequences originating from seen sources, the intra-source similarity is distributed with a mean of 0.932. 
In contrast, the inter-source similarity distribution is centered around 0.032.
For sequences from unseen sources, the intra-source similarity distribution has a mean of 0.893. 
The corresponding inter-source similarity distribution is centered around 0.019.
This indicates that the model captures source-specific information despite being trained in an unsupervised manner.
While this evaluation does not demonstrate semantic understanding, it provides evidence that contrastive learning results in a model that captures properties associated with individual sources.

\subsection{Consistency with Annotations}

Beyond identifying individual source IP addresses, we further investigate whether the learned representations relate to the sparse scanner network annotations introduced in Section \ref{sec:methodology:datasets}.

For this experiment, we draw 50.000 sequences labeled as belonging to a scanner network from \textit{Test-Unseen-Seq} and another 50.000 sequences labeled as belonging to a scanner network from \textit{Test-Unseen-Src}. 
For a quantitative analysis, we compare the cosine similarity of the learned representations of pairs of sequences from the same scanner network with that of pairs of sequences from distinct scanner networks.
Within the same scanner network, we exclude pairs of sequences originating from the same source IP address.
This is in order to focus on similarities different from the source identity already examined in Section \ref*{section:experiments-representation-consistency}. 

It can be seen in Figure~\ref{figure:scanner-cos-sim} that sequences associated with the same scanner network (intra) exhibit higher cosine similarity than sequences from distinct scanner networks (inter).
For unseen sequences originating from seen sources, the intra-class similarity distribution has a mean of 0.527. 
The corresponding inter-class similarity distribution is centered around -0.044.
For sequences from unseen sources, the intra-class similarity distribution has a mean of 0.529. 
The corresponding inter-class similarity distribution is centered around -0.043.

While the separation is less pronounced than for relating sources (Section~\ref{section:experiments-representation-consistency}), a shift in the similarity distributions can still clearly be observed.
Of note is the bimodal structure of the intra-class cosine similarity distribution. 
One mode is concentrated around high similarity values close to 1, while a second mode appears at substantially lower similarity values and overlaps with the inter-class distribution.
One possible explanation for this behavior is that scanner network annotations do not necessarily correspond to fully homogeneous groups. 
While IP addresses associated with the same scanner infrastructure may share certain properties, they may still exhibit different scanning strategies, target selections, or temporal behavior.
In addition, the lower-similarity mode of the intra-class distribution overlaps with the inter-class similarity distribution. 
This observation suggests that the provided scanner network annotations do not form perfectly separable subsets.
This could be due to sources belonging to different scanner networks performing similar scanning activities. 
E.g., distinct scanner infrastructures may employ similar scanning strategies, target selection schemes, or tooling, resulting in comparable sequential patterns.
In total, these findings indicate that contrastive learning induces semantically meaningful structure beyond simple memorization of individual source identities.
It is possible (but not established here) that the learned representations capture finer-grained distinctions even within the same scanner networks as well as similarities across different scanner networks that are not encoded in the annotation.

\subsection{Correlation Clustering}
\label{section:experiments-clustering}

In order to examine whether the learned representations can be used to algorithmically extract meaningful clusters of semantically similar sequences, we apply correlation clustering as described in Section~\ref{section:models-clustering}.

For this experiment, we consider three subsets, each containing 10.000 sequences drawn uniformly at random, without replacement, from \textit{Train}, \textit{Test-Unseen-Seq}, and \textit{Test-Unseen-Src}, respectively.
To better reflect realistic conditions in which scanner-related traffic is intermixed with unrelated traffic, the subsets are not restricted to sequences with scanner annotations. 
Instead, 50\% of the sequences are un-annotated, meaning that no scanner attribution is available for these.
The un-annotated sequences are clustered together with the annotated sequences and can therefore influence the result.

In order to quantify the accuracy of this result, the clustering, we observe for every pair of sequences whether these are in the same or distinct clusters, and calculate the precision, recall, accuracy and weighted F1-score of these binary decisions, only for those pairs in which both sequences are annotated.
Ideally, any two sequences with the same annotations would end up in the same cluster, and any two sequences with distinct annotations would end up in distinct clusters.
Unlabeled sequences are included in the clustering problem and can affect the result but are not considered in the calculation of metrics.
For correlation clustering, it is essential to learn the offset $\delta$ in (\ref{eq:clustering-costs}).
We do so by clustering the subset of sequences from \textit{Train} for every $\delta \in [0,1]$ with increments of $0.01$.
We fix $\delta' = 0.26$, which maximizes the weighted F1-score on \textit{Train} to 0.938.

With that $\delta'$, correlation clustering of the \textit{Test-Unseen-Seq} and \textit{Test-Unseen-Src} subsets have weighted F1-scores of 0.913 and 0.917, respectively.
Precision, recall, F1-scores for both cuts and joins, overall accuracy and weighted F1-score are reported in Table~\ref{tab:model_comparison}.
The computation time for greedy additive edge contraction is about 90 seconds on a single core of a i9-12900KF CPU.

These results indicate that the learned representations can be used to extract meaningful clusters of semantically similar sequences, and that the clustering generalizes to unseen data, especially to unseen sequences from sources.

Additionally, for exploratory analysis, we show two-dimensional t-SNE projections~\cite{tsne} of the learned embedding representations, in Figure~\ref{figure:tsne_embeddings_unseen}.
\vfill

\input{figures/src_ip_cos_sim.tex}
\input{figures/scanner_cos_sim.tex}
\input{figures/clustering_metrics.tex}

\begin{table}[ht]
    \small
    \centering    
    \begin{tabular}{lcccccccc}
    \toprule
    & \multicolumn{3}{c}{Cuts} & \multicolumn{3}{c}{Joins} & & \\
    \cmidrule(lr){2-4} \cmidrule(lr){5-7}
    Data 
    & Precision & Recall & F1 
    & Precision & Recall & F1 
    & Accuracy & Weighted F1 \\
    \midrule
    \textit{Test-Unseen-Seq} & 0.925 & 0.965 & 0.947 & 0.880 & 0.768 & 0.820 & 0.915 & 0.913 \\
    \textit{Test-Unseen-Src} & 0.949 & 0.948 & 0.948 & 0.799 & 0.802 & 0.800 & 0.918 & 0.917 \\
    \bottomrule
    \end{tabular}\\[1ex]
    \caption{Reported above are precision, recall, F1-score for both cuts and joins, as well as overall accuracy and weighted F1-score, for correlation clustering results on the \textit{Test-Unseen-Seq} and \textit{Test-Unseen-Src} subsets.}
    \label{tab:model_comparison}
\end{table}

\input{figures/clustering_tsne_vis_unseen_src.tex}

%% file: figures/src_ip_cos_sim.tex
\begin{figure}
	\centering
	\small
        \begin{tikzpicture}
        \begin{groupplot}[
        group style={
            group size=2 by 1,
            horizontal sep=1cm,
            vertical sep=2cm
            },
            width=6cm,
            height=4cm,
            xlabel={Cos Similarity},
            ylabel={Relative Frequency},
            grid=major
            ]
            \nextgroupplot[title={Unseen Sequences},legend to name={SrcLegend},legend style={legend columns=1},cells={anchor=west},legend image post style={xscale=0.5}]]
            \addplot[color=orange,line width=1pt]table[col sep=comma,x=bin_center,y=intra_density] {data/unseen_sequences_src_IP_all_similarity_histogram.csv};
            \addplot[color=teal,line width=1pt]table[col sep=comma,x=bin_center,y=inter_density] {data/unseen_sequences_src_IP_all_similarity_histogram.csv};
            \legend{intra,inter}
            \nextgroupplot[title={Unseen Sources},ylabel={}]
            \addplot[color=orange,line width=1pt]table[col sep=comma,x=bin_center,y=intra_density] {data/unseen_sources_src_IP_all_similarity_histogram.csv};
            \addplot[color=teal,line width=1pt]table[col sep=comma,x=bin_center,y=inter_density] {data/unseen_sources_src_IP_all_similarity_histogram.csv};
        \end{groupplot}
        \node[anchor=north west] at (group c1r1.north west){\ref{SrcLegend}};
    \end{tikzpicture}
	\caption{Shown above are distributions of cosine similarities of learned representations, for pairs of sequences from the same source IP (intra) and different source IP addresses (inter). 
	The analysis is performed separately for unseen sequences of seen sources (left), and for unseen sequences of unseen sources (right). 
	In both cases, a clear separation between the intra-source and inter-source similarity distributions can be observed, indicating that the learned model captures source-specific information.}
	\label{figure:scr-cos-sim}
\end{figure}
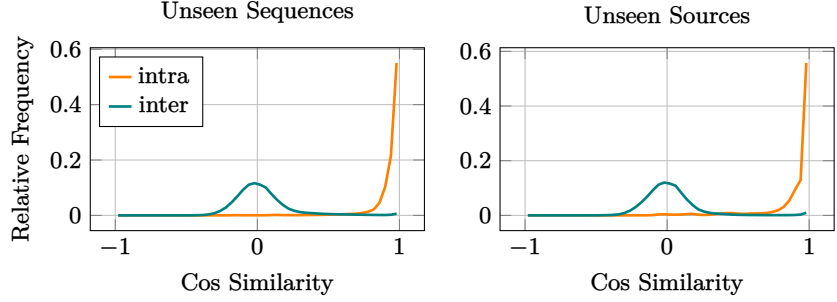

%% file: figures/scanner_cos_sim.tex
\begin{figure}
	\centering
	\small
        \begin{tikzpicture}
        \begin{groupplot}[
        group style={
            group size=2 by 1,
            horizontal sep=1cm,
            vertical sep=2cm
            },
            width=6cm,
            height=4cm,
            xlabel={Cosine Similarity},
            ylabel={Relative Frequency},
            grid=major,
            scaled y ticks=false,
            yticklabel style={
                /pgf/number format/fixed,
                /pgf/number format/precision=2
            },
            ]
            \nextgroupplot[title={Unseen Sequences},legend to name={ScannerLegend},legend style={legend columns=1},cells={anchor=west},legend image post style={xscale=0.5}]]
            \addplot[color=orange,line width=1pt]table[col sep=comma,x=bin_center,y=intra_density] {data/unseen_sequences_of_scanners_scanner_IDS_similarity_histogram.csv};
            \addplot[color=teal,line width=1pt]table[col sep=comma,x=bin_center,y=inter_density] {data/unseen_sequences_of_scanners_scanner_IDS_similarity_histogram.csv};
            \legend{intra,inter}
            \nextgroupplot[title={Unseen Sources},ylabel={}]
            \addplot[color=orange,line width=1pt]table[col sep=comma,x=bin_center,y=intra_density] {data/unseen_sources_of_scanners_scanner_IDS_similarity_histogram.csv};
            \addplot[color=teal,line width=1pt]table[col sep=comma,x=bin_center,y=inter_density] {data/unseen_sources_of_scanners_scanner_IDS_similarity_histogram.csv};
        \end{groupplot}
        \node[anchor=north west] at (group c1r1.north west){\ref{ScannerLegend}};
    \end{tikzpicture}
	\caption{
		Shown above are distributions of cosine similarities of learned representations, 
		for pairs of sequences annotated as belonging to the same scanner network (intra) and distinct scanner networks (inter). 
		The analysis is performed separately for unseen sequences of seen sources (left), and for unseen sequences of unseen sources (right). 
        In both cases, a clear separation can be observed. 
        The overlap of the distributions might be caused by annotations not corresponding to homogeneous categories.}
	\label{figure:scanner-cos-sim}
\end{figure}
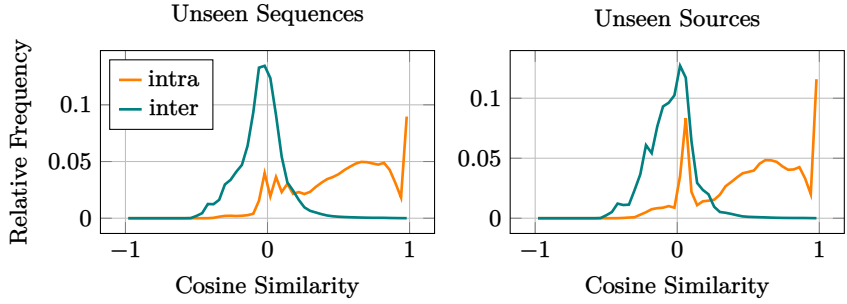

%% file: figures/clustering_metrics.tex
\begin{figure}
	\centering
	\small
    \begin{tikzpicture}
        \begin{groupplot}[
        group style={
            group size=2 by 1,
            horizontal sep=1cm,
            vertical sep=2cm
            },
            width=6cm,
            height=4cm,
            xlabel={Offset parameter $\delta$},
            ylabel={$F_1$ score},
            grid=major
            ]
            \nextgroupplot[title={Unseen Sequences},legend to name={ClusterLegend},legend style={legend columns=1,cells={anchor=west},legend image post style={xscale=0.5}}]]
            \addplot[color=orange,line width=1pt]table[col sep=comma,x=offset,y=f1_join] {data/unseen_sequences_of_scanners_scanner_ids_correlation_clustering_edge_metrics.csv};
            \addplot[color=teal,line width=1pt]table[col sep=comma,x=offset,y=f1_cut] {data/unseen_sequences_of_scanners_scanner_ids_correlation_clustering_edge_metrics.csv};
            \addplot[color=green,line width=1pt]table[col sep=comma,x=offset,y=f1_weighted] {data/unseen_sequences_of_scanners_scanner_ids_correlation_clustering_edge_metrics.csv};
            \draw[black,dashed,thick](axis cs:0.26,\pgfkeysvalueof{/pgfplots/ymin})--(axis cs:0.26,\pgfkeysvalueof{/pgfplots/ymax});
            \legend{join,cut,weighted}

            \nextgroupplot[title={Unseen Sources},ylabel={}]
            \addplot[color=orange,line width=1pt]table[col sep=comma,x=offset,y=f1_join] {data/unseen_sources_of_scanners_scanner_ids_correlation_clustering_edge_metrics.csv};
            \addplot[color=teal,line width=1pt]table[col sep=comma,x=offset,y=f1_cut] {data/unseen_sources_of_scanners_scanner_ids_correlation_clustering_edge_metrics.csv};
            \addplot[color=green,line width=1pt]table[col sep=comma,x=offset,y=f1_weighted] {data/unseen_sources_of_scanners_scanner_ids_correlation_clustering_edge_metrics.csv};
            \draw[black,dashed,thick](axis cs:0.26,\pgfkeysvalueof{/pgfplots/ymin})--(axis cs:0.26,\pgfkeysvalueof{/pgfplots/ymax});
        \end{groupplot}
        \node[anchor=south west] at (group c1r1.south west){\ref{ClusterLegend}};
    \end{tikzpicture}
	\caption{Shown above are the F1-scores of correlation clustering decisions for different values of the offset parameter $\delta$.
	These are shown for unseen sequences of network flow records from seen sources (left) and unseen sources (right). 
	The vertical dashed lines indicate $\delta' = 0.26$, which is optimal on the training set and used for evaluation.}
	\label{figure:clustering-curves}
\end{figure}
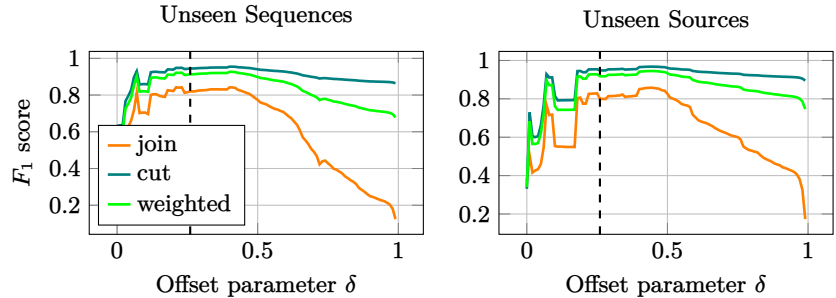

%% file: figures/clustering_tsne_vis_unseen_src.tex
\pgfplotsset{
    colormap={classes}{
        rgb255(0cm)=(220,220,220)   
        rgb255(1.9cm)=(100,225,25)
        rgb255(2cm)=(0,130,200)     
        rgb255(3cm)=(245,130,48)    
        rgb255(3.5cm)=(145,30,180)
        rgb255(4cm)=(60,180,75)     
        rgb255(4.1cm)=(230,25,75)
        rgb255(6.9cm)=(145,30,180)
        rgb255(7cm)=(0,255,0)    
        rgb255(7.5cm)=(0,0,255)
        rgb255(8cm)=(240,50,230)
    }
}

\begin{figure}
	\centering
	\small
        \begin{tikzpicture}
        \begin{groupplot}[
        group style={
            group size=2 by 2,
            horizontal sep=1cm,
            vertical sep=2cm
            },
            width=7cm,
            height=7cm,
            xlabel={$t$-SNE 1},
            ylabel={$t$-SNE 2},
            xmin=-110, xmax=110,
            ymin=-110, ymax=110,
            colormap name=classes
            ]
            \nextgroupplot[title={Scanner Labels \textit{Test-Unseen-Seq}},legend to name={CommonLegend},legend style={legend columns=3}]]
            \addplot[scatter,only marks,mark=*,mark size=0.4pt,scatter src=explicit,point meta=explicit,opacity=0.7,draw opacity=0,]table[col sep=comma,x=embedding_1,y=embedding_2,meta=label] {data/unseen_sequences_of_scanners_scanner_ids_tsne_visualization.csv};
            \nextgroupplot[title={Scanner Labels \textit{Test-Unseen-Src}},ylabel={}]
            \addplot[scatter,only marks,mark=*,mark size=0.4pt,scatter src=explicit,point meta=explicit,opacity=0.7,draw opacity=0]table[col sep=comma,x=embedding_1,y=embedding_2,meta=label] {data/unseen_sources_of_scanners_scanner_ids_tsne_visualization.csv};
        \end{groupplot}
        \path (group c1r1.south east) -- node[below=1cm]{\ref{CommonLegend}} (group c2r1.south west);
    \end{tikzpicture}
	\caption{Two-dimensional t-SNE projections of the embedded representations for 10,000 sequences sampled from \textit{Test-Unseen-Seq}, shown in the left panel, and 10,000 sequences sampled from \textit{Test-Unseen-Src}, shown in the right panel, as used in Section~\ref{section:experiments-clustering}. Colors indicate scanner network labels, with label 0 depicted in gray.}
	\label{figure:tsne_embeddings_unseen}
\end{figure}
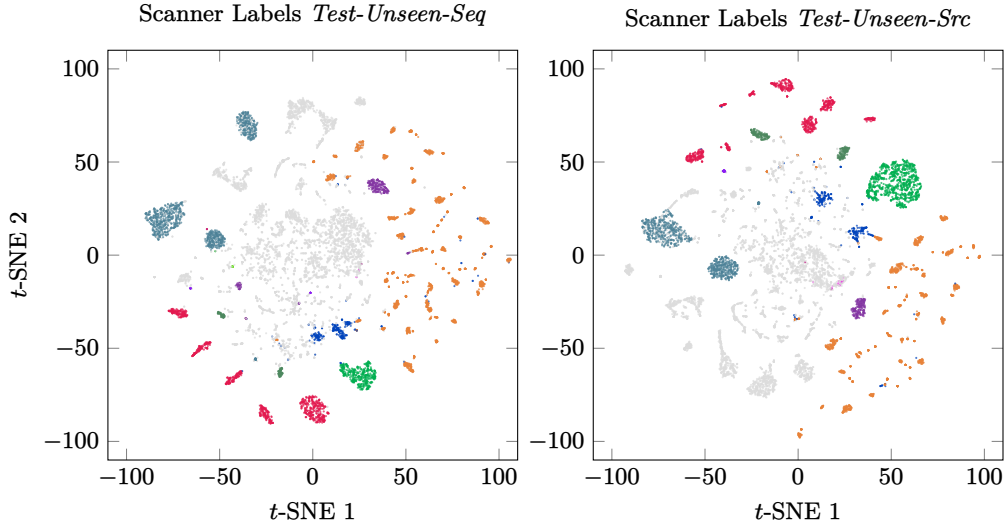

%% file: conclusion.tex
\section{Conclusion}

We explore whether semantically meaningful pairwise relationships between sequences of network flow records can be estimated by contrastive learning, without pretraining and without annotations.
The results suggest that embeddings can be obtained which encode meaningful source-specific information.
Despite the absence of semantic supervision, the learned embeddings exhibit scanner-related structure.
The observed overlap between intra-class and inter-class cosine similarity distributions suggests that the embedding space reflects additional structure not fully represented by the scanner annotations considered in this study.
Correlation clustering with respect to these similarities and a global threshold optimized on training data can be used to recover meaningful semantic groups under realistic conditions where scanner-related traffic is intermixed with unlabeled background traffic.
Overall, these results suggest that contrastive learning on minimally processed sequences of network flow records is capable of capturing semantically meaningful structure with potential applicability for exploratory traffic analysis and the investigation of previously unknown or weakly labeled network behavior.
At the same time, the presented results remain exploratory in nature. 
The scanner annotations used in this study are sparse and coarse-grained, and further work is required to better understand the semantic structure represented within the embedding space. 
Future research will focus on more extensive quantitative evaluation, analysis of alternative representation learning objectives, and means which can further improve the learned representations.

%% file: egbib.bib
@String(ICCV  = {Int. Conf. Comput. Vis.})

@String(NeurIPS = {Adv. Neural Inform. Process. Syst.})

@String(ICML  = {Int. Conf. Mach. Learn.})

@String(ICASSP=	{Int. Conf. on Acoustics, Speech and Sig. Proc.})

@String(ICCV  = {ICCV})

@String(NeurIPS = {NeurIPS})

@String(ICML  = {ICML})

@String(ICASSP=	{ICASSP})

@inproceedings{mikolov2013word2vec,
	author = {Mikolov, Tomas and Sutskever, Ilya and Chen, Kai and Corrado, Greg S and Dean, Jeff},
	booktitle = {NeurIPS},
	title = {Distributed Representations of Words and Phrases and their Compositionality},
	url = {https://proceedings.neurips.cc/paper_files/paper/2013/file/9aa42b31882ec039965f3c4923ce901b-Paper.pdf},
	year = {2013}
}

@inproceedings{dante,
	author = {Cohen, Dvir and Mirsky, Yisroel and Kamp, Manuel and Martin, Tobias and Elovici, Yuval and Puzis, Rami and Shabtai, Asaf},
	title = {DANTE: A Framework for Mining and Monitoring Darknet Traffic},
	year = {2020},
	booktitle = {25th European Symposium on Research in Computer Security (ESORICS)},
	doi = {10.1007/978-3-030-58951-6_5},
}

@INPROCEEDINGS{IP2Vec,
  author={Ring, Markus and Dallmann, Alexander and Landes, Dieter and Hotho, Andreas},
  booktitle={International Conference on Data Mining Workshops (ICDMW)}, 
  title={{IP2Vec: Learning Similarities Between IP Addresses}}, 
  year={2017},
  doi={10.1109/ICDMW.2017.93}
}

@inproceedings{DarkVec,
	author = {Gioacchini, Luca and Vassio, Luca and Mellia, Marco and Drago, Idilio and Houidi, Zied Ben and Rossi, Dario},
	title = {DarkVec: automatic analysis of darknet traffic with word embeddings},
	year = {2021},
	doi = {10.1145/3485983.3494863},
	booktitle = {International Conference on Emerging Networking EXperiments and Technologies (CoNEXT)},
}

@article{iDarkVec,
	author = {Gioacchini, Luca and Vassio, Luca and Mellia, Marco and Drago, Idilio and Houidi, Zied Ben and Rossi, Dario},
	title = {{i-DarkVec: Incremental Embeddings for Darknet Traffic Analysis}},
	journal = {ACM Trans. Internet Technol.},
	year = {2023},
	volume = {23},
	number = {3},
	doi = {10.1145/3595378},
}

@inproceedings{SimCLR,
	author = {Chen, Ting and Kornblith, Simon and Norouzi, Mohammad and Hinton, Geoffrey},
	title = {A simple framework for contrastive learning of visual representations},
	year = {2020},
	booktitle = {ICML},
	url = {https://proceedings.mlr.press/v119/chen20j/chen20j.pdf},
}

@inproceedings{koukoulis2025SelfSupTransformerIDS,
  author    = {Ippokratis Koukoulis and Ilias Syrigos and Thanasis Korakis},
  title     = {Self-Supervised Transformer-based Contrastive Learning for Intrusion Detection Systems},
  booktitle = {Proceedings of the IFIP Networking 2025 Conference},
  year      = {2025},
  publisher = {IFIP},
  address   = {Limassol, Cyprus},
  url       = {https://networking.ifip.org/2025/images/Net25_papers/1571127827.pdf}
}

@article{FlowTransformer,
	author = {Manocchio, Liam Daly and Layeghy, Siamak and Lo, Wai Weng and Kulatilleke, Gayan K. and Sarhan, Mohanad and Portmann, Marius},
	title = {{FlowTransformer: A transformer framework for flow-based network intrusion detection systems}},
	journal = {Expert Syst. Appl.},
	year = {2024},
	volume = {241},
	number = {C},
	doi = {10.1016/j.eswa.2023.122564},
}

@inproceedings{NetGlyph,
	author = {Noblet, Gabin and Lefebvre, Cédric and Owezarski, Philippe and Ritchie, William},
	title = {{NetGlyph: Representation Learning to generate Network Traffic with 	Transformers}},
	booktitle = {International Conference on Network and Service Management (CNSM)},
	year = {2024},
	doi = {10.23919/CNSM62983.2024.10814626}
}

@misc{NetGPT,
      title={{NetGPT: Generative Pretrained Transformer for Network Traffic}}, 
      author={Xuying Meng and Chungang Lin and Yequan Wang and Yujun Zhang},
      year={2025},
      eprint={2304.09513},
      archivePrefix={arXiv},
      primaryClass={cs.NI},
      url={https://arxiv.org/abs/2304.09513}, 
}

@INPROCEEDINGS{PNetGPT,
  author={Li, Zedong and Li, Shijie and Fang, Dongliang and Chen, Xin and Song, Zhanwei and Li, Zhi and Lv, Shichao and Sun, Limin},
  booktitle={International Conference on Acoustics, Speech and Signal Processing (ICASSP)}, 
  title={{PNetGPT: Proprietary Protocol Network Traffic Generation with Pre-trained Transformer}}, 
  year={2025},
  doi={10.1109/ICASSP49660.2025.10890383}
}

@article{GPTOnTheWire,
	author = {Delgado-Soto, Javier Aday and L\'{o}pez de Vergara, Jorge E. and Gonz\'{a}lez, Iv\'{a}n and Perdices, Daniel and de Pedro, Luis},
	title = {{GPT on the wire: Towards realistic network traffic conversations generated with large language models}},
	journal = {Comput. Netw.},
	year = {2025},
	volume = {265},
	number = {C},
	doi = {10.1016/j.comnet.2025.111308},
}

@inproceedings{BERT,
    title = "{BERT}: Pre-training of Deep Bidirectional Transformers for Language Understanding",
    author = "Devlin, Jacob and Chang, Ming-Wei and Lee, Kenton and Toutanova, Kristina",
    booktitle = "Proceedings of the 2019 Conference of the North {A}merican Chapter of the Association for Computational Linguistics: Human Language Technologies, Volume 1 (Long and Short Papers)",
    publisher = "Association for Computational Linguistics",
    doi = "10.18653/v1/N19-1423",
    year={2019},
}

@INPROCEEDINGS{TRWPortscanDetection,
  author={Jaeyeon Jung and Paxson, V. and Berger, A.W. and Balakrishnan, H.},
  booktitle={IEEE Symposium on Security and Privacy}, 
  title={Fast portscan detection using sequential hypothesis testing}, 
  year={2004},
  doi={10.1109/SECPRI.2004.1301325}
}

@INPROCEEDINGS{LimitationsTRW,
  author={Mell, Peter and Harang, Richard},
  booktitle={Conference on Communications and Network Security (CNS)}, 
  title={Limitations to threshold random walk scan detection and mitigating enhancements}, 
  year={2013},
  doi={10.1109/CNS.2013.6682723}
}

@article{EvasionResistantScanDetection,
	author = {Harang, Richard and Mell, Peter},
	year = {2015},
	volume = {4},
	number = {4},
	pages = {1--10},
	title = {Evasion-resistant network scan detection},
	journal = {Security Informatics},
	doi = {10.1186/s13388-015-0019-7}
}

@article{GraphBasedBotnetDetection,
	author = {Chowdhury, Sudipta and Zhang, Fangyan and Zhang, Song and Medal, Hugh and Marufuzzaman, Mohammad and Bian, Linkan},
	year = {2017},
	pages = {14},
	title = {Botnet detection using graph-based feature clustering},
	volume = {4},
	journal = {Journal of Big Data},
	doi = {10.1186/s40537-017-0074-7}
}

@inproceedings{GAEC,
  author    = {Margret Keuper and Evgeny Levinkov and Nicolas Bonneel
               and Guillaume Lavou{\'e} and Thomas Brox and Bjoern Andres},
  title     = {Efficient Decomposition of Image and Mesh Graphs by Lifted Multicuts},
  booktitle = {ICCV},
  year      = {2015},
  doi = {10.1109/ICCV.2015.204}
}

@INPROCEEDINGS{DynamicClusterAnalysisTelescopes,
	author={Huang, Kai and Gioacchini, Luca and Mellia, Marco and Vassio, Luca},
	booktitle={European Symposium on Security and Privacy Workshops (EuroS \& PW)}, 
	title={Dynamic Cluster Analysis to Detect and Track Novelty in Network Telescopes}, 
	year={2024},
	doi={10.1109/EuroSPW61312.2024.00037}
}

@ARTICLE{EmbeddingsTransferForTrafficAnalysis,
  author={Gioacchini, Luca and Mellia, Marco and Vassio, Luca and Drago, Idilio and Milan, Giulia and Houidi, Zied Ben and Rossi, Dario},
  journal={IEEE Transactions on Network and Service Management}, 
  title={Cross-Network Embeddings Transfer for Traffic Analysis}, 
  year={2024},
  volume={21},
  number={3},
  pages={2686-2699},
  doi={10.1109/TNSM.2023.3329442}
}

@article{HDBSCAN,
	author = {Campello, Ricardo J. G. B. and Moulavi, Davoud and Zimek, Arthur and Sander, J\"{o}rg},
	title = {Hierarchical Density Estimates for Data Clustering, Visualization, and Outlier Detection},
	year = {2015},
	publisher = {Association for Computing Machinery},
	volume = {10},
	number = {1},
	doi = {10.1145/2733381},
	journal = {ACM Trans. Knowl. Discov. Data},
}

@InProceedings{scrcd-iidad-22,
	author    = {Sommese, Raffaele and Claffy, KC and van Rijswijk-Deij, Roland and Chattopadhyay, Arnab and Dainotti, Alberto and Sperotto, Anna and Jonker, Mattijs},
	booktitle = {Internet Measurement Conference (IMC)},
	title     = {{Investigating the Impact of DDoS Attacks on DNS Infrastructure}},
	year      = {2022},
	doi       = {10.1145/3517745.3561458},
}

@InProceedings{jkkrs-mtuam-17,
	author    = {Jonker, Mattijs and King, Alistair and Krupp, Johannes and Rossow, Christian and Sperotto, Anna and Dainotti, Alberto},
	booktitle = {Internet Measurement Conference (IMC)},
	title     = {{Millions of Targets under Attack: A Macroscopic Characterization of the DoS Ecosystem}},
	year      = {2017},
	doi       = {10.1145/3131365.3131383},
}

@INPROCEEDINGS{bazn-sldda-14,
	author={Balkanli, Eray and Alves, Jander and Zincir-Heywood, A. Nur},
	booktitle={Symposium on Computational Intelligence in Cyber Security (CICS)},
	title={{Supervised Learning to detect DDoS Attacks}},
	year={2014},
	doi={10.1109/CICYBS.2014.7013367},
}

@InProceedings{bzh-fsrbd-15,
  	author= {Eray Balkanli and Nur Zincir{-}Heywood and Malcolm I. Heywood},	
	booktitle = {40th Local Computer Networks Conference Workshops (LCNW)},
	title     = {{Feature Selection for Robust Backscatter DDoS Detection}},
	year      = {2015},
	doi       = {10.1109/LCNW.2015.7365905},
}

@article{dsacc-acioc-14,
	author={Dainotti, Alberto and Squarcella, Claudio and Aben, Emile and Claffy, Kimberly C. and Chiesa, Marco and Russo, Michele and Pescapé, Antonio},
	title={{Analysis of Country-Wide Internet Outages Caused by Censorship}}, 
	journal={IEEE/ACM Transactions on Networking}, 
	publisher={IEEE},
	year={2014},
	volume={22},
	number={6},
	pages={1964--1977},
	doi={10.1109/TNET.2013.2291244}
}

@InProceedings{pfxrm-mvicm-21,
	author    = {Padmanabhan, Ramakrishna and Filast\`{o}, Arturo and Xynou, Maria and Raman, Ram Sundara and Middleton, Kennedy and Zhang, Mingwei and Madory, Doug and Roberts, Molly and Dainotti, Alberto},
	booktitle = {ACM SIGCOMM Workshop on Free and Open Communications on the Internet (FOCI)},
	title     = {{A Multi-Perspective View of Internet Censorship in Myanmar}},
	year      = {2021},
	doi       = {10.1145/3473604.3474562},
}

@Article{mpsss-isw-03,
	author  = {David Moore and
                  Vern Paxson and
                  Stefan Savage and
                  Colleen Shannon and
                  Stuart Staniford{-}Chen and
                  Nicholas Weaver},
	journal = {{IEEE} Security {\&} Privacy},
	title   = {{Inside the Slammer Worm}},
	year    = {2003},
	number  = {4},
	pages   = {33--39},
	volume  = {1},
	doi     = {10.1109/msecp.2003.1219056},
}

@Article{dkcpp-assfb-15,
	author  = {Alberto Dainotti and Alistair King and Kimberly Claffy and Ferdinando Papale and Antonio Pescape},
	journal = {{IEEE}/{ACM} Transactions on Networking},
	title   = {{Analysis of a {\textquotedblleft}/0{\textquotedblright} Stealth Scan From a Botnet}},
	year    = {2015},
	number  = {2},
	pages   = {341--354},
	volume  = {23},
	doi     = {10.1109/tnet.2013.2297678},
	publisher={IEEE},
	address={Piscataway, NJ, USA},
}

@InProceedings{aabbb-umb-17,
	author    = {Antonakakis, Manos and April, Tim and Bailey, Michael and Bernhard, Matthew and Bursztein, Elie and Cochran, Jaime and Durumeric, Zakir and Halderman, J. Alex and Invernizzi, Luca and Kallitsis, Michalis and Kumar, Deepak and Lever, Chaz and Ma, Zane and Mason, Joshua and Menscher, Damian and Seaman, Chad and Sullivan, Nick and Thomas, Kurt and Zhou, Yi},
	booktitle = {26th USENIX Security Symposium (USENIX)},
	title     = {{Understanding the Mirai Botnet}},
	year      = {2017},
	doi= {10.5555/3241189.3241275}
}

@InProceedings{pkb-hdpdt-22,
	author    = {Morteza Safaei Pour and Joseph Khoury and Elias Bou-Harb},
	booktitle = {{IEEE}/{IFIP} Network Operations and Management Symposium (NOMS)},
	title     = {{{HoneyComb}: A Darknet-Centric Proactive Deception Technique For Curating {IoT} Malware Forensic Artifacts}},
	year      = {2022},
	doi       = {10.1109/noms54207.2022.9789827},
}

@Article{kphwy-dicsb-22,
	author  = {Michalis Kallitsis and Rupesh Prajapati and Vasant Honavar and Dinghao Wu and John Yen},
	journal = {{IEEE} Transactions on Information Forensics and Security},
	title   = {{Detecting and Interpreting Changes in Scanning Behavior in Large Network Telescopes}},
	address      = {Piscataway, NJ, USA},
	publisher = {IEEE},
	year    = {2022},
	pages   = {3611--3625},
	volume  = {17},
	doi     = {10.1109/tifs.2022.3211644},
}

@inproceedings {mvs-iidsa-01,
	author = {David Moore and Geoffrey M. Voelker and Stefan Savage},
	title = {{Inferring Internet {Denial-of-Service} Activity}},
	booktitle = {USENIX Security Symposium},
	year = {2001},
}

@misc{ucsd-stardust,
	author={CAIDA},
	year={2021},
	title={{Sustainable Tools for Analysis and Research on Darknet Unsolicited Traffic}},
	url={https://www.caida.org/projects/stardust/},
}

@inproceedings{hnbkh-adeci-24,
	author = {Raphael Hiesgen and Marcin Nawrocki and Marinho Barcellos and Daniel Kopp and Oliver Hohlfeld and Echo Chan and Roland Dobbins and Christian Doerr and Christian Rossow and Daniel R. Thomas and Mattijs Jonker and Ricky Mok and Xiapu Luo and John Kristoff  and  Thomas C. Schmidt and Matthias W{\"a}hlisch and KC Claffy},
	title = {{The Age of DDoScovery: An Empirical Comparison of Industry and Academic DDoS Assessments}},
	booktitle = {Internet Measurement Conference (IMC)},
	year = {2024},
	doi          = {10.1145/3646547.3688451},

}

@inproceedings{mmcgm-lloln-25,
  author = {Alexander M{\"a}nnel and  Jonas M{\"u}cke and kc Claffy and Max Gao and Ricky K. P. Mok and Marcin Nawrocki and Thomas C. Schmidt and Matthias W{\"a}hlisch},
  title = {{Lessons Learned from Operating a Large Network Telescope}},
  booktitle = {Proc. of ACM Special Interest Group on Data Communication (SIGCOMM)},
  pages = {826--841},
  year = {2025},
  publisher = {ACM},
  address = {New York, NY, USA},
  doi = {10.1145/3718958.3754347},
}

@misc{c-as-21,
  author       = {Michael Collins},
  title = {{Acknowledged Scanners}},
  url          = {https://gitlab.com/mcollins_at_isi/acknowledged_scanners.git},
  year         = {2021}
}

@misc{c-cis-17,
  author       = {{Censys}},
  howpublished = {Website},
  title        = {{Censys Plattform}},
  year         = {2017},
  url          = {https://platform.censys.io/},
}

@misc{s-swfse-14,
  author       = {{Shodan}},
  howpublished = {{Website}},
  title        = {{Shodan - the world's first search engine for Internet-connected devices.}},
  url          = {https://www.shodan.io/},
}

@article{mnhsz-wqpmu-25,
  author = {Jonas M{\"u}cke and Marcin Nawrocki and Raphael Hiesgen and Patrick Sattler and Johannes Zirngibl and Georg Carle and Jan Luxemburk and Thomas C. Schmidt and Matthias W{\"a}hlisch},
  title = {{Waiting for QUIC: Passive Measurements to Understand QUIC Deployments}},
  journal = {Proceedings of the ACM on Networking (PACMNET)},
  pages = {41:1-41:26},
  volume = {3},
  number = {CoNEXT4},
  year = {2025},
  publisher = {ACM},
  address = {New York},
  url   = {https://doi.org/10.1145/3768988},
}

@misc{caida-url-1,
	url = {https://www.caida.org/about/legal/aua/},
}

@misc{caida-url-2,
	url = {https://www.caida.org/catalog/datasets/request_user_info_forms/telescope_dataset_request/},
}

@misc{git-url,
	url = {https://github.com/JannikPresberger/Contrastive_Learning_and_Correlation_Clustering_for_Sequences_of_Network_Telescope_Data},
}

@article{chopra-1993,
    author = {Chopra, Sunil and Rao, M. R.},
    year = {1993},
    title = {The partition problem},
    journal = {Mathematical Programming},
    volume = {59},
    number = {1},
    pages = {87--115},
    doi = {10.1007/BF01581239},
}

@article{tsne,
  author  = {Laurens van der Maaten and Geoffrey Hinton},
  title   = {Visualizing Data using t-SNE},
  journal = {Journal of Machine Learning Research},
  year    = {2008},
  volume  = {9},
  number  = {86},
  pages   = {2579--2605},
  url     = {http://jmlr.org/papers/v9/vandermaaten08a.html}
}
